\documentclass[11pt]{article}
\usepackage{times} % Make text more compressed.
\usepackage{url} % Allow URLs.
\usepackage{graphicx} % Allow inclusion of graphics.
\usepackage[margin=1in]{geometry} % Sets page size and margins
\usepackage{authblk} % Put authors names in a block.
\usepackage{amsmath} % Allow fancy math stuff
\usepackage{amsfonts}
\usepackage{tabularx,ragged2e,booktabs} % Stuff for tables.
\usepackage{titlesec}
\usepackage{amssymb}
\usepackage{comment}
\usepackage{subcaption}
\usepackage{authblk}
\usepackage{xcolor} % Allow colors.
\usepackage{hyperref}
\usepackage{amssymb}
\usepackage{dsfont}
\usepackage{float}
\usepackage[ruled,vlined]{algorithm2e}
\usepackage{multirow}
\usepackage[normalem]{ulem}
\usepackage{graphicx,lipsum,wrapfig}% http://ctan.org/pkg/{graphicx,lipsum,wrapfig}
\usepackage{cleveref}
\usepackage[flushleft]{threeparttable}

\usepackage{booktabs,caption}
\usepackage[sorting=none,style=ieee,citestyle=numeric-comp, backend=biber]{biblatex}
\bibliography{bib}

\usepackage{wrapfig}
\usepackage{color}

\begin{document}

\author[1]{Michal Golovanevsky}
\author[1,2,*]{Carsten Eickhoff}
\author[1,3,*]{Ritambhara Singh}

\affil[1]{Department of Computer Science, Brown University}
\affil[2]{Center for Biomedical Informatics, Brown University}
\affil[3]{Center for Computational Molecular Biology, Brown University}
\affil[*]{Co-corresponding Authors}
\date{\vspace{-1.6cm}}

\title{Multimodal Attention-based Deep Learning for Alzheimer's Disease Diagnosis}

\maketitle

\vspace{0.2cm}
\begin{abstract}
\noindent Alzheimer's Disease (AD) is the most common neurodegenerative disorder with one of the most complex pathogeneses, making effective and clinically actionable decision support difficult. The objective of this study was to develop a novel multimodal deep learning framework to aid medical professionals in AD diagnosis.
We present a Multimodal Alzheimer's Disease Diagnosis framework (MADDi) to accurately detect the presence of AD and mild cognitive impairment (MCI) from imaging, genetic, and clinical data. MADDi is novel in that we use cross-modal attention, which captures interactions between modalities - a method not previously explored in this domain. We perform multi-class classification, a challenging task considering the strong similarities between MCI and AD. We compare with previous state-of-the-art models, evaluate the importance of attention, and examine the contribution of each modality to the model's performance.
MADDi classifies MCI, AD, and controls with 96.88\% accuracy on a held-out test set. When examining the contribution of different attention schemes, we found that the combination of cross-modal attention with self-attention performed the best, and no attention layers in the model performed the worst, with a 7.9\% difference in F1-Scores.
Our experiments underlined the importance of structured clinical data to help machine learning models contextualize and interpret the remaining modalities. Extensive ablation studies showed that any multimodal mixture of input features without access to structured clinical information suffered marked performance losses.
This study demonstrates the merit of combining multiple input modalities via cross-modal attention to deliver highly accurate AD diagnostic decision support.

\end{abstract}

\noindent
\textbf{Available at:} \url{https://github.com/rsinghlab/MADDi}

\section{Introduction}
\subsection{Background and Significance}

Alzheimer's Disease (AD) is the most common neurodegenerative disorder affecting approximately 5.5 million people in the United States and 44 million people worldwide \cite{naqvi_2017}. Despite extensive research and advances in clinical practice, less than 50\% of AD patients are diagnosed accurately for pathology and disease progression based on clinical symptoms alone\cite{thies_bleiler_2013}. The pathology of the disease occurs several years before the onset of clinical symptoms, making the disease difficult to detect at an early stage \cite{iddi_li_aisen_rafii_thompson_donohue_2019}. Mild cognitive impairment (MCI) is considered an AD prodromal phase, where the gradual change from MCI to AD can take years to decades \cite{petersen_2016}. As AD cannot currently be cured, but only delayed in progression, early detection of MCI before irreversible brain damage occurs is crucial for preventive care. 

The urgent need for clinical advancement in AD diagnosis inspired the Alzheimer’s Disease Neuroimaging Initiative (ADNI) to collect diverse data such as imaging, biological markers, and clinical assessment on MCI and AD patients \cite{mueller2005ways}. Such distinct data inputs are often referred to as individual \textit{modalities}; a research problem is characterized as \textit{multimodal} when it considers multiple such modalities and \textit{unimodal} when it includes just one. Thanks to data collection efforts such as the one spearheaded by ADNI, it became possible to create unimodal machine learning models capable of aiding AD diagnosis, most commonly using imaging data \cite{forouzannezhad2020gaussian, uysal2020hippocampal, dimitriadis2018random, beheshti2017classification, dyrba2015predicting}, or clinical records \cite{el2021alzheimer, zhou2013modeling}. Recently, deep learning (DL) has shown considerable potential for clinical decision support and outperformed traditional unimodal machine learning methods in AD detection \cite{WANG2019145, UYSAL2020108669, KRUTHIKA201934}. The major strength of DL over traditional machine learning models is the ability to process large numbers of parameters and effectively learn meaningful connections between features and labels. Even with DL's advantage, single-modality input is often insufficient to support clinical decision-making \cite{stahlschmidt2022multimodal}. 
  
Alzheimer's Disease diagnosis is multimodal in nature - health care providers examine patient records, neurological exams, genetic history, and various imaging scans. Integrating multiple such inputs provides a more comprehensive view of the disease. Thus, several deep learning-based multimodal studies \cite{ELSAPPAGH2020197,ABUHMED2021106688,BUCHOLC2019157,venugopalan2021multimodal} have leveraged the richer information encoded in multimodal data. Despite an overall convincing performance, they all miss a crucial component of multimodal learning - cross-modal interactions. The existing methods simply concatenate features extracted from the different modalities to combine their information, limiting the model's ability to learn a shared representation \cite{ngiam2011multimodal}. In response, we propose a novel Multimodal Alzheimer's Disease Diagnosis framework (MADDi), which uses a cross-modal attention scheme \cite{Tan2019LXMERTLC} to integrate imaging (magnetic resonance imaging (MRI)), genetic (single nucleotide polymorphisms (SNPs)), and structured clinical data to classify patients into control (CN), MCI, and AD groups. 

Many successful studies were conducted using the ADNI dataset \cite{mueller2005ways}. Only a small subset of them used multimodal data, and an even smaller subset attempted three-class classification. In this work, we propose to use attention as a vehicle for cross-modality interactions. We show state-of-the-art performance on the challenging multimodal three-class classification task, achieving 96.88\% average test accuracy across 5 random model initializations. Next, we investigated the contribution of each modality to the overall model. While for unimodal models, images produced the most robust results (92.28\% accuracy), when we combined all three data inputs, we found that the clinical modality complements learning the best. Lastly, since our method utilizes two different types of neural network attention, we investigated the contribution of each type and found significant performance improvements when using attention layers over no attention. Through our experiments, we were able to highlight the importance of capturing interactions between modalities.

\section{Methods and Materials}
\subsection{Data Description}

The data used in this study were obtained from the ADNI database (\url{https://adni.loni.usc.edu/}), which provides imaging, clinical, and genetic data for over 2220 patients spanning four studies (ADNI1, ADNI2, ADNI GO, and ADNI3). The primary goal of ADNI has been to test whether combining such data can help measure the progression of MCI and AD. Our study follows the common practice of using patient information from only ADNI1, 2, and GO, since ADNI 3 is still an ongoing study expected to conclude in 2022. To capture a diverse range of modalities, we focused on patients with imaging, genetic, and clinical data available. We trained unimodal models on the full number of participants per modality. For our multimodal architecture, we only used those patients that had all three modalities recorded (referred to as the \textit{overlap dataset}). The number of participants in each group is detailed in Table \ref{tab:participants}.

\begin{table}[H] 
\centering
\caption{\textbf{Number of participants in each modality and their diagnosis.} This table shows the number of participants in each modality and further separates the participants into their diagnoses. The Overlap section refers to patients that had all three modalities recorded.}
\begin{tabular}{|l|l|l|l|l|} 
\hline
         & \textbf{Total Participants} & \textbf{Control} & \textbf{Mild Cognitive Impairment} & \textbf{Alzheimer's Disease} \\ \hline
Clinical & 2384               & 942     & 796                         & 646                 \\ \hline
Imaging  & 551                & 278     & 123                         & 150                 \\ \hline
Genetic  & 805                & 241     & 318                         & 246                 \\ \hline
Overlap  & 239                & 165     & 39                          & 35                  \\ \hline
\end{tabular}
%\begin{tablenotes}
%  \small
%  \item 
%\end{tablenotes}
\label{tab:participants}
\end{table}

\subsubsection{Ground Truth Labels}
Since ADNI's launch in 2003, more participants have been added to each phase of the study, and the disease progression of initial participants is continuously followed. Over time, some patients that were initially labeled as control (CN) and mild cognitive impairment (MCI) had a change in diagnosis as their disease progressed. While some patients had as many as 16 MRI scans since the start of the study, clinical evaluations were collected much less frequently, and genetic testing was only performed once per patient. Thus, combining three modalities per patient was a unique challenge as, at times, there were contradictory diagnoses, making the ground truth diagnosis unclear. For our overlap dataset, we used the latest MRI and clinical evaluation for each patient and the most recent diagnosis. Several studies focused on using time-series data to track the progression of the disease \cite{ELSAPPAGH2020197,ABUHMED2021106688,guerrero2016instantiated,jedynak2012computational,yau2015longitudinal}. However, we aimed to accurately classify patients into groups at the most recent state of evaluation so our method can be generalized to patients who are not part of long-term observational studies. 

\subsubsection{Clinical Data Pre-processing}
For clinical data, we use 2384 patients' data from the neurological exams (e.g., balance test), cognitive assessments (e.g., memory tests), and patient demographics (e.g., age). The clinical data is quantitative, categorical, or binary, totaling 29 features. We removed any feature that could encode direct indication of AD (e.g., medication that a patient takes to manage extant AD). A full list of features can be found in Supplement \ref{section:clinical}. Categorical data were converted to features using one-hot encoding, and continuous-valued features were normalized. 

\subsubsection{Genetic Data Pre-processing}
\label{subsubsection:genetic_proc}
The genetic data consists of the whole genome sequencing (WGS) data from 805 ADNI participants by Illumina’s non-Clinical Laboratory Improvement Amendments (non-CLIA). The resulting variant call files (VCFs) were generated by ADNI using Burrows-Wheeler Aligner and Genome Analysis Toolkit in 2014. Each subject has about 3 million SNPs in the raw VCF file generated. However, not all detected SNPs are informative in predicting AD. We followed the established pre-processing steps detailed in \cite{venugopalan2021multimodal} to reduce the number of SNPs and keep only the relevant genetic components. After such filtering (detailed further in Supplement \ref{section:genetic}), we had 547,863 SNPs per patient. As we only have 805 patients with genetic data, we were left with a highly sparse matrix. We used a Random Forest classifier as a supervised feature selection method to determine which are the most important features, reducing our feature space to roughly 15,000 SNPs. Note that data points used for model testing were not seen by the classifier. While the result was still sparse, we found that this level was a reasonable stopping point as determined by the experiment detailed in Supplementary \ref{section:genetic}. The final data were grouped into three categories: no alleles, any one allele, or both alleles present.

\subsubsection{Imaging Data Pre-processing}
\begin{figure}[h!]
\begin{center}
  \caption{\textbf{Imaging input example.} The imaging model took as input three slices from the center of the MRI brain volume, that were uniformly shaped to 72 x 72 pixels.}
  \label{fig:mri}
  \includegraphics[width=8cm,]{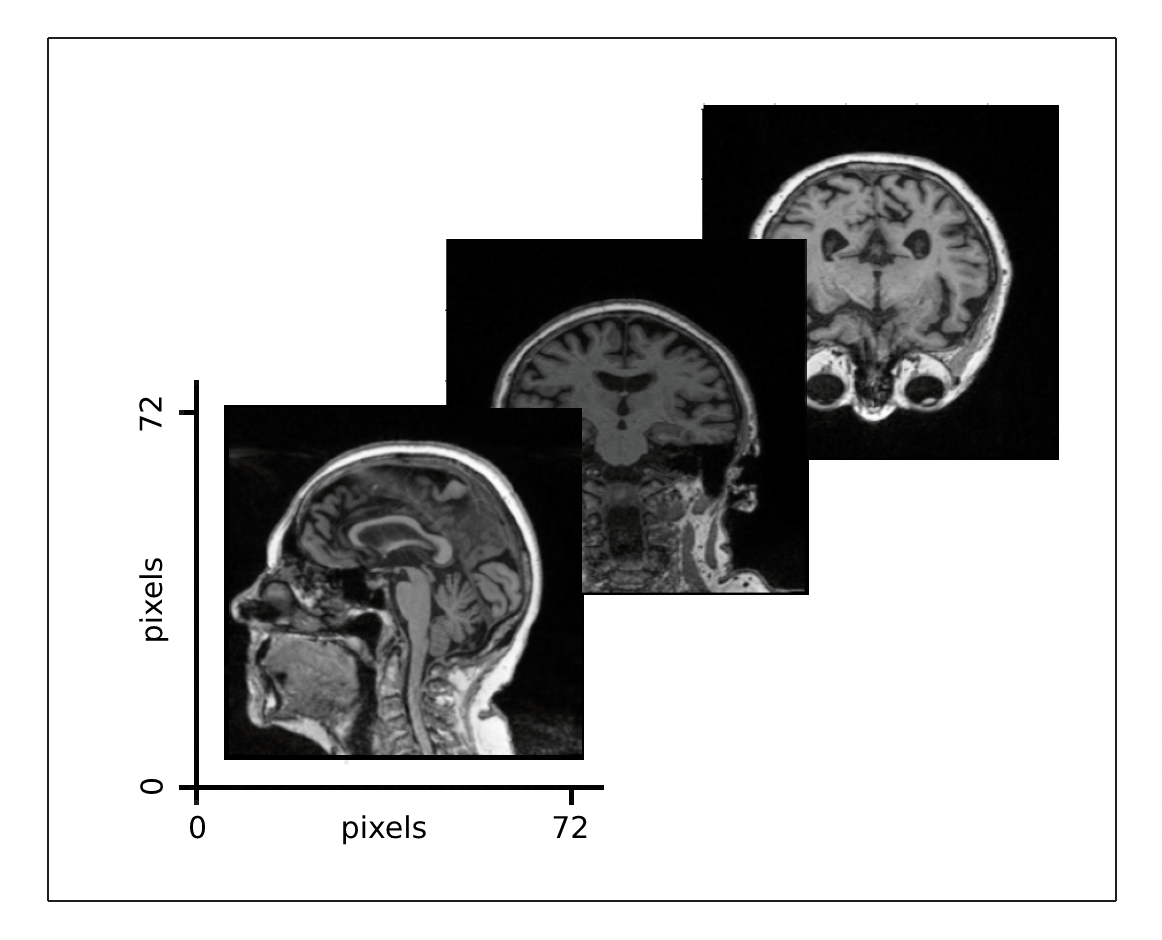}
 \end{center}
\end{figure}

The imaging data in this study consist of cross-sectional MRI data corresponding to the first baseline screenings from ADNI1 (551 patients). The data publisher has standardized the images to eliminate the non-linearities caused by the scanners from different vendors. From each brain volume, we used three slices corresponding to the center of the brain in each dimension. An example input is shown in Figure \ref{fig:mri}. Further details on the ADNI pre-processing steps and experiments justifying the use of three images per patient can be found in Supplement \ref{section:img_preproc}.

\subsubsection{Finalizing the Training Dataset}
To train our multimodal architecture, we used 239 patients that had data available from all three modalities. The overlap dataset was chosen out of the original data mentioned above - imaging (551 patients), SNP (805 patients), and clinical (2284 patients) to predict AD stages. While the SNP data was unique per patient, the clinical and imaging data appeared multiple times. To ensure a proper merger, we used the timestamps in the clinical data and matched it to the closest MRI scan date. Next, we used the most recent evaluation on a patient to avoid repeating patients. The patients' demographic information is shown in Table~\ref{tab:demographic}.

\begin{table}[H]
\centering
\caption{\textbf{Participants' demographic information.} This table shows the number of participants in each diagnosis group along with the percent of females and the average age of each group.}
\begin{tabular}{|l|l|l|l|}
\hline
\textbf{Group}                         & \textbf{Participants (n)} & \textbf{Female Sex (\%)} & \textbf{Mean Age (years)} \\ \hline
Control                       & 165              & 53.9            & 77.8             \\ \hline
Mild Cognitive Impairment & 39               & 34.2            & 76.6             \\ \hline
Alzheimer's Disease           & 35               & 31.4            & 78.1             \\ \hline
\end{tabular}
%\begin{tablenotes}
%  \small
%  \item   
%\end{tablenotes}
\label{tab:demographic} 
\end{table}

\subsection{Multimodal Framework}
\label{section:mm_frame}
\begin{figure}[H]
\begin{center}
    \caption{\textbf{Model Architecture.} (a) Data inputs - clinical data (demographics, memory tests, balance score, etc.), genetic (SNPs), and imaging (MRI scans). (b) The input sources are combined and fed into a fully connected (FC) neural network architecture for genetic and clinical modalities and a convolutional neural network (CNN) for imaging data. (c) Using the obtained features from the neural networks, a self-attention layer reinforces any inner-modal connections. (d) Then, each modality pair is fed to a bi-directional cross-modal attention layer which captures the interactions between modalities. (e) Lastly, the outputs are concatenated and passed into a decision layer for classification into the (f) output Alzheimer’s stages (CN, MCI, and AD).}
  \includegraphics[scale=0.35]{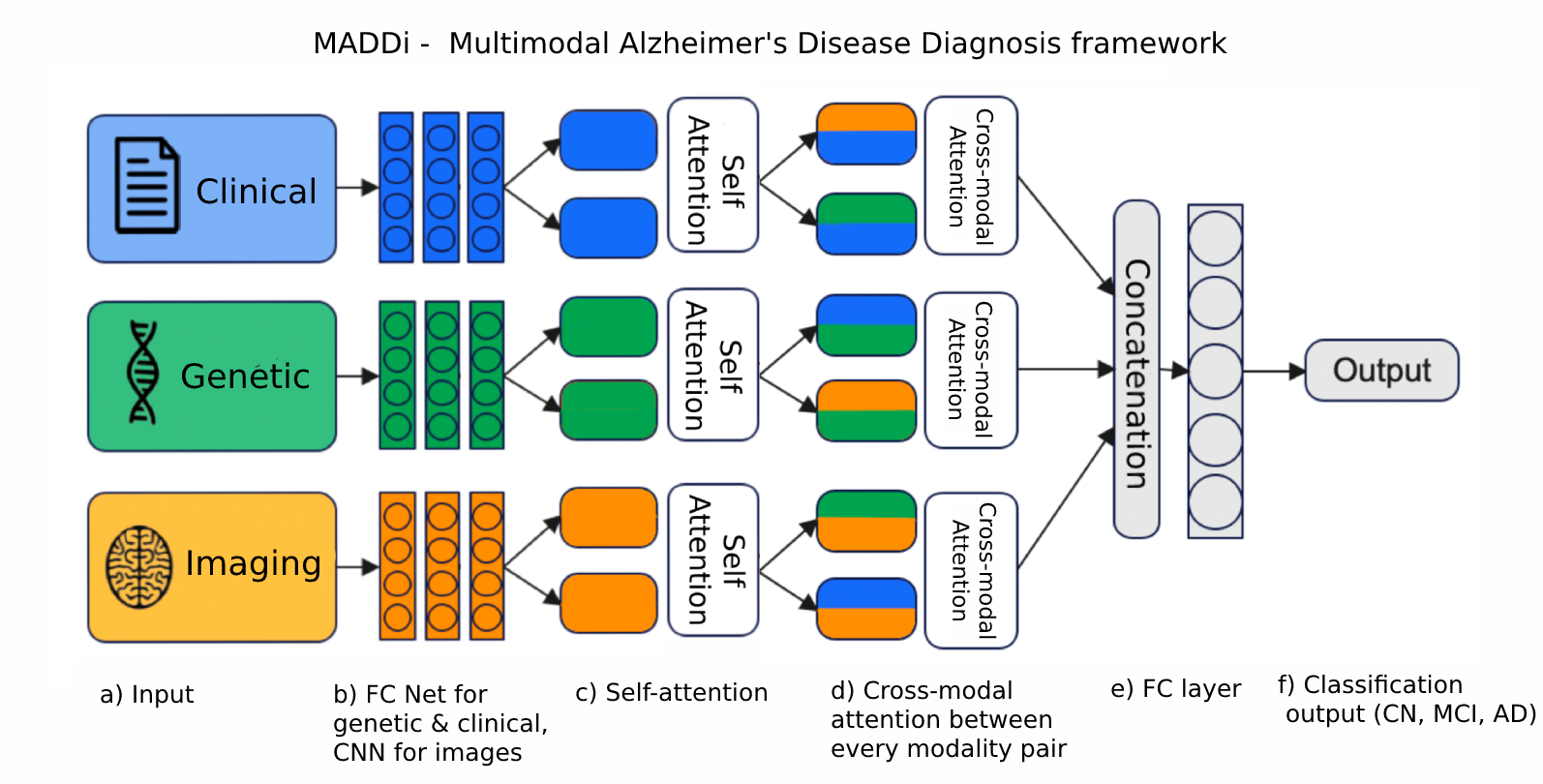}
  \label{fig:arch}
  \end{center}

\end{figure}

The proposed framework, MADDi, is shown in Figure \ref{fig:arch}. The model receives a patient's pre-processed clinical, genetic, and imaging data and outputs the corresponding diagnosis (CN, AD, or MCI). Following the input, there are modality-specific neural network architecture backbones developed in the single modality setting (further detailed in the {\hyperref[subsection:uni_frame]{Performance of Unimodal Models} Section)}. For clinical and genetic data, this is a three-layer fully connected network, and for imaging data, we have a three-layer convolutional neural network. The output of those layers then enters a multi-headed self-attention layer, which allows the inputs to interact with each other and find what features should be paid most attention to within each modality. This layer is followed by a cross-modal bi-directional attention layer \cite{Tan2019LXMERTLC}, which performs a similar calculation to self-attention but across different pairs of modalities. The purpose of cross-modal attention is to model an interaction between modalities; for example, clinical features may help reinforce what the imaging features tell the model and thus lead to more robust decision making. Both attention types are rigorously defined in the {\hyperref[section:att]{Neural Network Attention} Section}. The last step concatenates the output of the parallel attention computations and feeds it into a final dense layer that makes the prediction. 

\subsection{Experimental Design}

\subsubsection{Neural Network Attention}
\label{section:att}

As a part of our experimental design, we evaluate the importance of attention in our model. Previous methods \cite{ELSAPPAGH2020197,ABUHMED2021106688, venugopalan2021multimodal} explored the value that DL brings to automating AD diagnosis. We build on top of previous multimodal DL frameworks and examine the need for inter-modal interactions through attention. Thus, we used the same framework but toggled the presence of attention based on four criteria: self-attention and cross-modal attention, just self-attention, just cross-modal attention, and no attention. The different types of attention are introduced in the following. 

\textit{Generalized attention}: Attention is a mechanism that captures the relationship between two states and highlights the features that contribute most to decision-making. An attention layer takes as input queries and keys of dimension $d_k$, and values of dimension $d_v$. A key is the label of a feature used to distinguish it from other features, and a query is what checks all available keys and selects the one that matches best. We compute the dot products of the query with all keys, divide each by the square root of $d_k$, and apply a Softmax function, which converts a vector of numbers into a vector of probabilities, to obtain the weights on the values:
\begin{equation}
    Attention(Q, K, V) = softmax(\frac{QK^T}{ \sqrt{d_k}})V
\end{equation}
Following the success of the Transformer \cite{Vaswani2017AttentionIA}, we use the multi-head attention module, which allows the model to jointly attend to information from different representation subspaces at different positions.

\textit{Self-Attention}: For self-attention mechanisms, queries, keys, and values are equal. The self-attention mechanism allows us to learn the interactions among the different parts of an input(``self'') and determine which parts of the input are relevant for making predictions (``attention''). In our case, the prior neural network layers produce in parallel three latent feature matrices for each modality that act as the keys, queries, and values: imaging matrix $I$, genetic matrix $G$, and clinical matrix $C$. Self-attention, in our terms, refers to attention computation done within the same modality. Thus the self-attention module performs the following operations:
\begin{align}
&\operatorname{self-attention}( I \rightarrow I)\\
&\operatorname{self-attention}( G \rightarrow G)\\
&\operatorname{self-attention}( C \rightarrow C)
\end{align}
  
\textit{Cross-modal attention}: In each bi-directional cross-modal attention layer \cite{Tan2019LXMERTLC}, there are two unidirectional cross-modal attention sub-layers: one from $modality\ 1$ to $modality\ 2$ and one from $modality\ 2$ to $modality\ 1$. In our case, the cross-modal attention layer takes the output of each self-attention computation: image self-attention output $I_s$, genetic self-attention output $G_s$, and clinical self-attention output $C_s$. Thus the cross-modal attention module performs the following operations:
  \begin{align}
    &concatenation(\operatorname{cross-modal\ attention}( I_s \rightarrow C_s), \operatorname{cross-modal \ attention}( C_s \rightarrow I_s))\\
    &concatenation(\operatorname{cross-modal \ attention}( C_s \rightarrow G_s), \operatorname{cross-modal \ attention}( G_s \rightarrow C_s))\\
    &concatenation(\operatorname{cross-modal \ attention}( G_s \rightarrow I_s), \operatorname{cross-modal \ attention}( I_s \rightarrow G_s))
  \end{align}
Lastly, we created a model with no attention module at all, where, following the neural network layers, we directly proceed to concatenate and produce output through the final dense layer. This setting represents the previous state-of-the-art methods used for integrating multimodal datasets for our task.

\subsubsection{Unified Hyperparameter Tuning Scheme}
The modality-specific neural network part of MADDi was predetermined based on the hyperparameter tuning done on each unimodal model (Supplement \ref{section:hyper}).We did not use the overlapping test set during hyperparameter tuning was done. To fairly evaluate the need for attention, we tuned using the same hyperparameter grid for each of the other experimental models. Meaning, that each model (self-attention only, the cross-modal attention only, and the model with no attention) gets its own set of best hyperparameters. 
We first randomly split our 239 patients into a training set (90\%) and a held-out testing set (10\%). We chose a 90 - 10 split for consistency with all the papers we compared against (shown in \ref{tab:comparison}). We designed a 3-fold cross-validation scheme and took the parameters that gave the best average cross-validation accuracy. Next, we used 5 random seeds to give the model multiple attempts at initialization. The best initialization was determined based on the best training performance on the full train and validations set (i.e., \ validation was added into training). This pipeline was repeated to find until we found the best parameters for each baseline. 

\subsubsection{Evaluation Metrics}
The following metrics were calculated for each model: accuracy, precision (positive predictive value), recall (sensitivity), and F1-score (harmonic mean of recall and precision). F1-score was the primary performance metric for evaluating our baselines. Accuracy was used to evaluate our best model against previous papers, as that was the metric most commonly reported on this task. The metric calculations are detailed in Supplement \ref{eval_formula}.

\section{Results}
\subsection{Performance of Unimodal Models}
\label{subsection:uni_frame}
To demonstrate the success of our multimodal framework, MADDi, we first examined the performance of a single-modality model. Our evaluation pipeline was consistent across all modalities in that we used a neural network and then tuned hyperparameters to find the best model. We split each modality into training (90\%) and testing (10\%) data, where the testing set was not seen until after the best parameters were chosen using the average accuracy across 3-fold cross-validation. The reported test accuracies are averaged across five random initializations, which remained consistent across all modalities. 
The results are summarized in Figure \ref{fig:uni} (and in Supplement \ref{eval_unimodal} Table \ref{tab:uni}). For the clinical unimodal model, we created a neural network with three fully connected layers (other hyperparameters can be found in Supplement \ref{section:hyper}). The best model yielded $80.5\%$ average accuracy across five random seeds. The model was trained on 2384 patients.
For imaging results, we created a convolutional neural network with three convolution layers. The best model yielded $92.28\%$ average accuracy across five random seeds. The model was trained on 551 patients, but we allowed for patient repetitions as we found that only using 551 images was not enough to train a successful model. Thus, we had 3674 MRI scans from 551 patients (some patients repeated up to 16 times). We selected our testing set such that it has 367 unique MRIs (10\% of training), and we do not allow for any repeating patients in the testing set. We only allowed repetition during training, and no patient was included in both training and testing sets. 
For genetic data, we created a neural network with three fully connected layers. The best model yielded $77.78\%$ average accuracy across five random seeds. The model was trained on 805 unique patients.

\begin{figure}[h!]
\begin{center} 
    \caption{\textbf{Metric evaluation of unimodal models.} The graph shows all four evaluation metrics for the best neural network model for each modality - imaging, clinical, and genetic. The imaging model gives the best performance overall, whereas the genetic modality gives the lowest performance with highest variation.}
    \label{fig:uni}
  \includegraphics[scale=0.5]{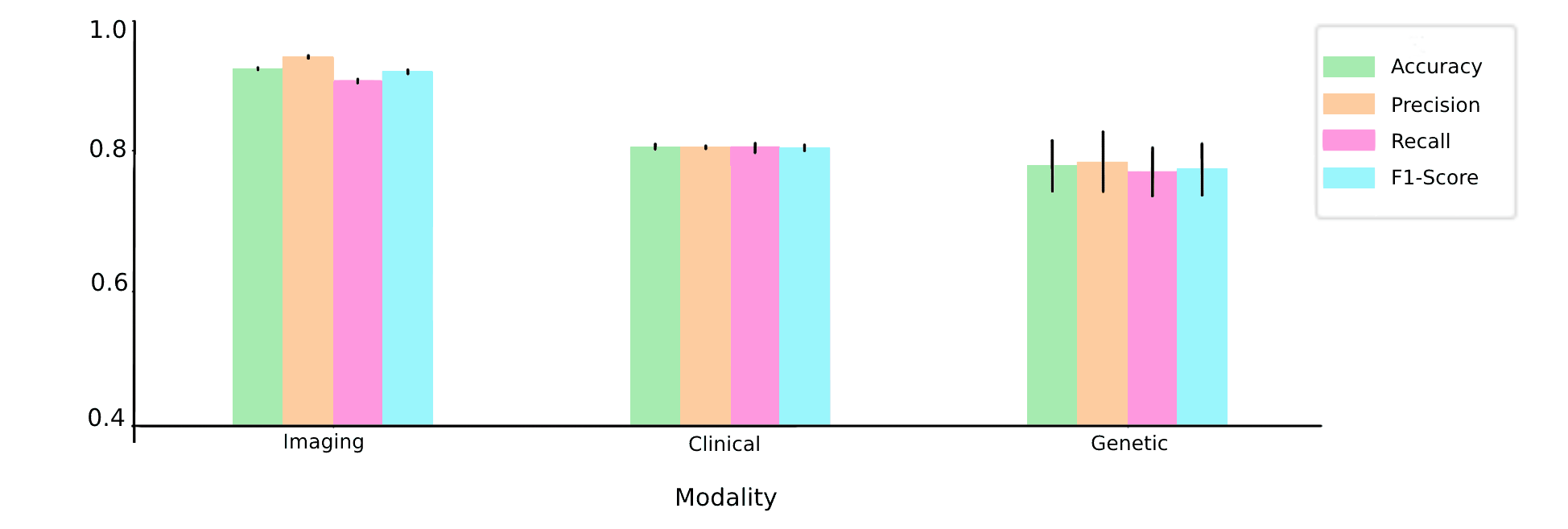}

  \end{center}
\end{figure}

\subsection{Performance of Multimodal Models}
Table \ref{tab:comparison} contrasts the performance and architecture of MADDi with state-of-the-art multimodal approaches from the literature. Note that due to the differences in dataset characteristics and multitask settings, it was not possible to directly compare performance among methods that only report binary classification or use a single modality. Thus, we only report studies that used two or more modalities and did 3-class classification. For our proposed method we report the average accuracy across 5 random initializations on a held-out test set. Therefore, we also use the test (as opposed to cross-validation) accuracy from other studies. 
Bucholc et al.\ \cite{BUCHOLC2019157} used support vector machines to classify patients into normal, questionable cognitive impairment (QCI), and mild/moderate AD, comparable to our definitions of control, MCI, and AD. They reported 82.9\% test accuracy but did not rely on deep learning.
Fang et al.\ \cite{FANG2020108856} used Gaussian discriminative component analysis as a method of multi-class classification using two different imaging modalities, achieving 66.29\% accuracy on the test set. 
Abuhmen et al.\ and El-Sappagh et al.\ \cite{ABUHMED2021106688, ELSAPPAGH2020197} both used MRI, PET, and various health record features. The key difference between the two is that El-Sappagh et al.\ considered a four-class classification of control, AD, stable MCI (patients who do not progress to AD), and progressive MCI. Since they did not report three-class classification, we could not directly compare to their work, but note that they achieved 92.62\% accuracy on the 4-class task. Both methods utilized deep learning, but they focused more on disease progression diagnosis with time-series data rather than static disease diagnosis. 
Venugopalan et al.\ \cite{venugopalan2021multimodal} were most similar to our study in structure, modality choice, and pre-processing. They, too, did not utilize the recent advancement of attention-based multimodal learning, which is where our architecture stands out. 
At $96.88\% \pm 3.33\%$ average accuracy, MADDi defined state-of-the-art performance on the multimodal three-class classification task.

\newcolumntype{s}{>{\columncolor{yellow}}c}
\begin{table}[H]
\begin{center}
\caption{\textbf{Comparison with related studies.} This table shows the comparison between our study and 5 other previous studies that attempted to solve a similar problem to ours. MADDi preformed with 96.88\% average accuracy and 91.41\% average F1-Score across 5 random initializations on a held-out test set, achieving state-of-the-art performance on the multimodal three-class classification task. }
\scalebox{0.8}{%
\begin{tabular}{ | l | l | l | l | l |}
\hline
\textbf{Study}                                          & \textbf{Modality}           & \textbf{Accuracy} & \textbf{F1-Score} & \textbf{Method}         \\ \hline
Bucholc et al.  \cite{BUCHOLC2019157}, 2019                & MRI, PET, Clinical          & 82.90\%           & Not Reported      & SVM                     \\ \hline
Fang et al.  \cite{FANG2020108856}, 2020                   & MRI, PET                    & 66.29\%           & Not Reported      & GDCA                    \\ \hline
Abuhmed et al.  \cite{ABUHMED2021106688}, 2021             & MRI, PET, Clinical          & 86.08\%           & 87.67\%           & Multivariate BiLSTM     \\ \hline
Venugopalan et al.  \cite{venugopalan2021multimodal}, 2021 & MRI, SNP, Clinical          & 78\%              & 78\%              & DL + RF                 \\ \hline
\textbf{MADDi}                                                             & \textbf{MRI, SNP, Clinical} & \textbf{96.88\%}  & \textbf{91.41\%}  & \textbf{DL + Attention} \\ \hline
\end{tabular}}
%\begin{tablenotes}
%  \small
%  \item This table shows the comparison between our study and 5 other previous studies that attempted to solve a similar problem to ours.    
%\end{tablenotes}
\label{tab:comparison}
\end{center}
\end{table}

\subsection{Model Robustness}
\label{subsection:robust}
To definitively conclude that both self-attention and cross-modal attention are necessary additions to the model, we ablated the attention schemes in three conditions (self-attention only, cross-modal attention only, and the model with no attention) on the held-out test set using the best parameters for each respective model. To demonstrate that our success was not dependent on initialization, we used 100 different random seeds and recorded the distribution of F1-scores on the testing set. For more information on our test sample selection, refer to Supplement \label{section:samp_sel}. Figure \ref{fig:attention} (and Table \ref{tab:attention} in Supplement Section \ref{tab_boxplot}) shows that self-attention and cross-modal attention layers together have the narrowest distribution, with the highest median F1-score. The next best distribution is the cross-modal attention layer alone, which has a slightly wider distribution but still the second-highest median F1-score. The success of the two methods involving cross-modal attention becomes apparent and provides strong evidence that capturing interactions between modalities positively influences the model’s decision-making. All three models that utilize attention achieved 100\% F1-score for at least one initialization, while the model with no attention layers only reached at most 92.8\% F1-score. Furthermore, the performance of our final model was 7.9\% average F1-Score higher than a model with no attention, and was significant ($p-value<0.0001$, two-sample Z-test) - providing further evidence that attention was beneficial for multimodal data integration.
\begin{figure}[H]
\begin{center}
    \caption{\textbf{F1-Score Distribution for different attention-based and attention-free baselines.} Box plots showing the F1-score distribution across 100 random seeds to demonstrate the value of attention in a deep learning model. The F1-scores were calculated from a held-out test set. The horizontal line represents the median F1-score, and the boxes represent the first and third quartiles. The whiskers represent quartile 1 - (1.5 x interquartile range) and quartile 3 + (1.5 x interquartile range). The dots represent the individual F1-scores for each model. ****: P $\le$ 0.0001. The figure demonstrates that the combination of self-attention with cross-modal attention performs the best with the most narrow variation. }
    
  \includegraphics[width=0.7\textwidth]{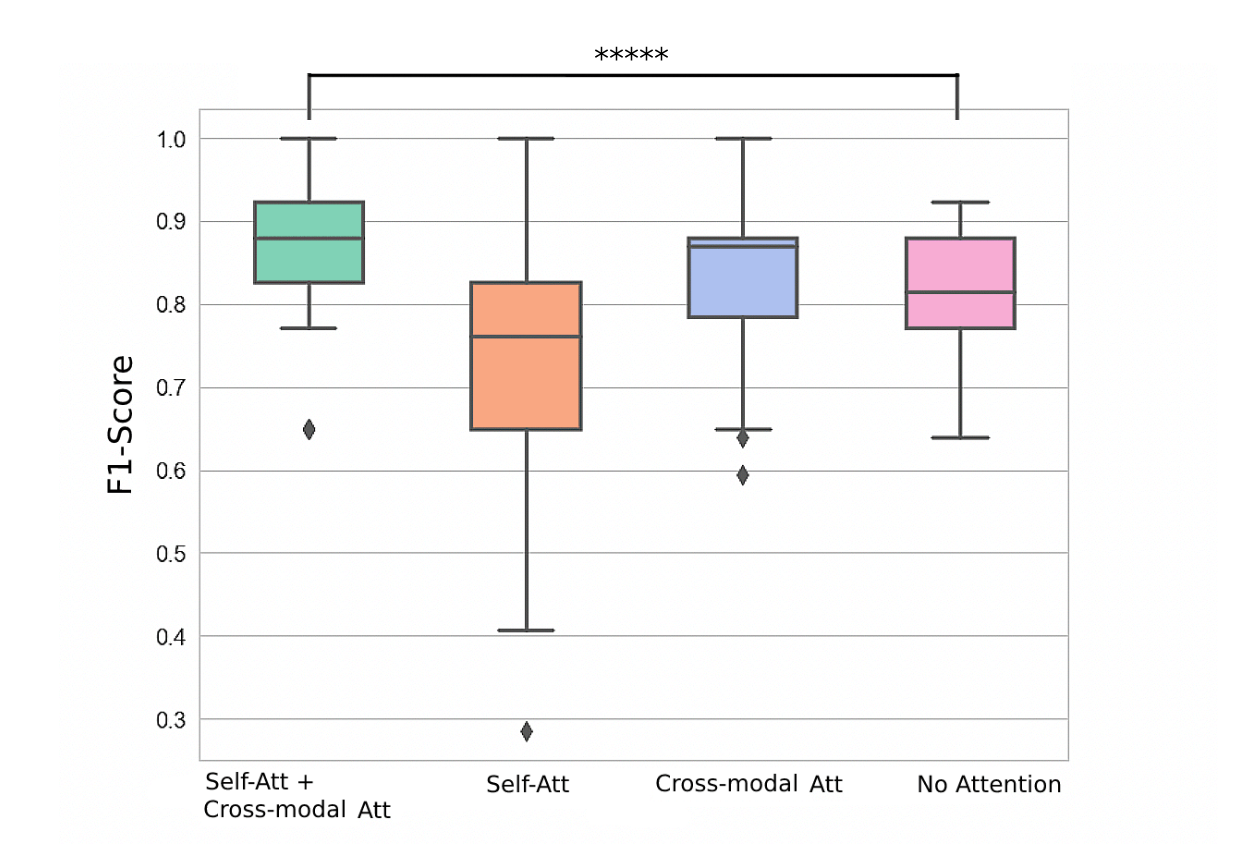}
  \label{fig:attention}
  \end{center}
\end{figure}

Using the self-attention and cross-attention model (MADDi), we investigated the performance of the model with respect to the individual classes as seen in Table \ref{tab:classes}. We report metrics averaged across 5 random initializations. We find that, regardless of the initialization,  the model is extremely accurate at identifying AD patients. However, for some cases, it tends to mistake MCI patients for controls. We hypothesize  that, since our data does not include different stages of MCI, it may have MCI patients with mild symptoms that could be mistaken for controls by the model. These observations can be seen in detail through 5 confusion matrices from the 5 initializations in the Supplement Section \ref{confusion} Figure \ref{fig:confusion}.

\begin{table}[H]
\begin{center}
\caption{\textbf{Investigating performance for each individual class.} This table shows the performance metrics averaged across 5 random initializations of MADDi on each class (control, Moderate Cognitive Impairment, and Alzheimer's Disease). We observe that Alzheimer's Disease is predicted correctly regardless of initialization and the only mistake the model makes is misclassifying MCI patients as control patients.}\label{tab:classes}
\begin{tabular}{|l|l|l|l|l|}
\hline
                              & \textbf{Accuracy} & \textbf{Precision} & \textbf{Recall} & \textbf{F1-Score} \\ \hline
Control                       & 96.66\%           & 96.78\%            & 98.88\%         & 97.81\%           \\ \hline
Moderate Cognitive Impairment & 96.66\%           & 90.00\%            & 70.00\%         & 76.66\%           \\ \hline
Alzheimer's Disease           & 100\%             & 100\%              & 100\%           & 100\%             \\ \hline
\end{tabular}
\end{center}
\end{table}

\subsection{Modality Importance}
\label{subsection:eval_imp}
\begin{figure}[h!]
\begin{center}
  \caption{\textbf{Evaluation of modality importance.} This figure evaluates the possible combinations of modalities. The metrics were calculated as an average of five random initializations on a held-out test set. The combination of the three modalities performs the best across all evaluation metrics. Excluding the clinical modality causes a drop in performance, demonstrating the value of clinical information.}
  \label{fig:eval_metric}
  \includegraphics[width=\textwidth]{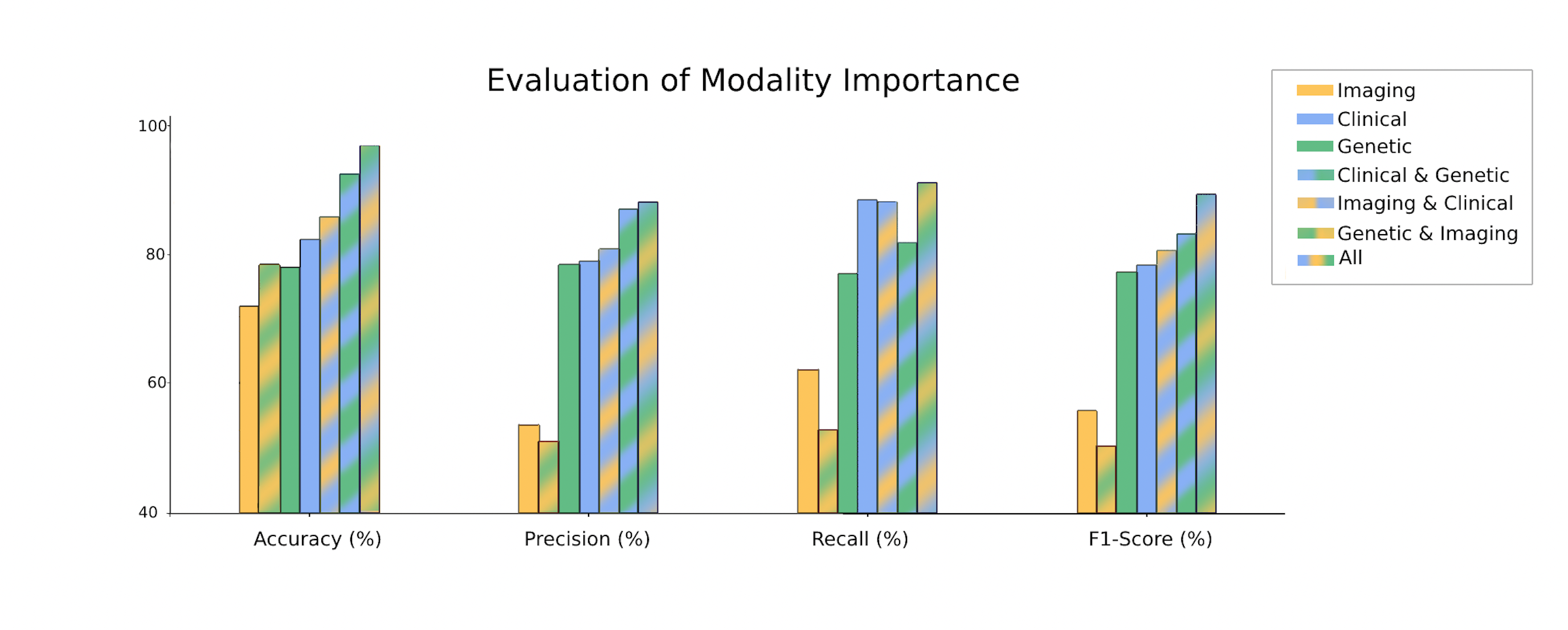}
\end{center}
\end{figure}

Finally, we investigated the importance of each modality to bring more transparency into the model’s decision-making and motivate future data collection efforts. Knowing how valuable each modality is to disease classification and what happens when it is excluded from the experiment can shed light on where to focus scientific resources. While every study participant had at least some clinical data available, only a few hundred patients had MRI scans. To fairly compare each modality's importance to the model, we performed our analyses on the same exact patients. Thus, we evaluated the contribution of the modalities on the overlap patient set (detailed in Figure \ref{fig:eval_metric}). For single modalities, we only used self-attention. For a pair of modalities, we used both self-attention and cross-modal attention. We performed hyperparameter tuning for each model to ensure fair evaluation, with all the parameters provided in Supplement \ref{section:hyper} Table \ref{tab:hyper}. 
As seen in Figure \ref{fig:eval_metric}, combining the three modalities performs the best across all evaluation metrics. A full table with the numeric results can be found in Supplement \ref{section:modality_importance} Table \ref{tab:modal_imp}. The interesting discovery here was the clinical modality contribution to this performance. While the use of two modalities was better than one in most cases, when clinical data was withheld, we saw a significant drop in performance; clinical data alone achieved 82.29\% accuracy and 78.30\% F1-score, whereas genetic and imaging together achieved 78.33\% accuracy and dropped to 50.07\% F1-score. These results suggest that the clinical dataset is an important catalyst modality for AD prediction. We hypothesize that this empirical merit stems from the fact that clinical features offer the necessary patient context that grounds the additional modalities such as vision or omics information and helps the model in their effective representation and interpretation.

To further investigate the clinical data, we used a Random Forest classifier (which fits the clinical training data to the diagnosis labels in a supervised manner) to find the most dominant features from the clinical modality: memory score, executive function score, and language score. A full list of features in order of importance can be found in Supplement \ref{section:clinical} Figure \ref{fig:clinical_imp}. 

\section{DISCUSSION}
\subsection{Clinical Importance and Implications}
Detecting AD accurately is clinically important as it is the 6th leading cause of death in the United States and is the most common cause of dementia \cite{naqvi_2017}. Without treatment or other changes to the trajectory, aggregate care costs for people with AD are projected to increase from \$203 billion in 2013 to \$1.2 trillion in 2050 \cite{thies_bleiler_2013}.
Despite studies such as ADNI collecting various imaging, genetic, and clinical data to improve our understanding of the underlying disease processes, most computational techniques still focus on using just a single modality to aid disease diagnosis. Our state-of-the-art model allows for an effective integration scheme of three modalities and can be expanded as more data sources become available. 

\subsection{Future Extensions}
The proposed model architecture can be used in other multimodal clinical applications, such as cancer detection \cite{esteva2017dermatologist, weng2017combining}. As the efforts to make health care data more broadly available continue to increase, we believe that our model will help aid the diagnostic process. The framework we propose does not rely on modality-specific processing within the model itself. Thus, our future work will include other data (PET scans, clinical notes, biomarkers, etc.). While it is straightforward to simply interchange the current modalities with new ones and only use three modalities at a time, we plan on expanding our work beyond this current level as there is no rigid constraint on the number of modalities used with the given codebase. 
Furthermore, similarly to the task El-Sappagh et al.\ \cite{ELSAPPAGH2020197} explored, we will extend our task to more than three-class classification and use our work to detect different types of MCI (stable MCI and progressive MCI). This will help better understand AD progression and delay the change from MCI to AD.

\subsection{Limitations}

When creating our unimodal performance baselines, we often struggled with finding the ground truth labels for the genetic data. While every patient had a diagnosis attached to an MRI scan, and most of the clinical exams also had a diagnosis listed, genetic data did not. Out of the 808 patients with genetic data, we used 805 patients where diagnosis on their most recent MRI scan agreed with their clinical diagnosis. Thus, we proceeded with 805 patients to eliminate any error in the ground truth labeling. This gap is natural, as a patient may have had a more recent MRI that changed the diagnosis, leaving the recent clinical evaluation outdated (and vice versa).

During pre-processing of the MRI images, we chose to use the middle slice of the brain rather than the full brain volume. This could mean that our model did not see certain areas of the brain. When running unimodal experiments on the MRI data, the performance remained the same (within one percent) when using just the middle slice of the brain compared to the full brain volume, shown in the Supplement \ref{section:img_preproc}. Since processing thousands of slices per patient is much more computationally expensive, we proceeded with this simplification. While on our task, there was no significant difference in performance; in other applications integrating the full brain volumes into the model could further increase performance.

\section{Conclusions}

In this work, we presented a Multimodal Alzheimer's Disease Diagnosis framework (MADDi), which uses attention-based deep learning to detect Alzheimer's disease. The performance of MADDi was superior to that of existing multimodal machine learning methods and was shown to be consistently high regardless of chance initialization. We offer three distinct contributions: integrating multimodal inputs, multi-task classification, and cross-modal attention for capturing interactions. Many existing multimodal DL models simply concatenate each modality's features despite substantial differences in feature spaces. We employed attention modules to address this problem; self-attention reinforces the most important features extracted from the neural network backbones, and cross-modality attention \cite{Tan2019LXMERTLC} reinforces relationships between modalities. Combining the two attention modules resulted in a 96.88\% accuracy and defined state-of-the-art on this task. Overall, we believe that our approach demonstrates the potential for an automated and accurate deep learning method for disease diagnosis. We hope that in the future, our work can be used to integrate multiple modalities in clinical settings and introduce the highly effective attention-based models to the medical community.

\section{Acknowledgements} 
We thank Alzheimer’s Disease Neuroimaging Initiative (ADNI) database (\url{https://adni.loni.usc.edu/}) for providing data for this study. For a complete list of ADNI investigators, refer to: \url{https://adni.loni.usc.edu/wp-content/uploads/how_to_apply/ADNI_Acknowledgement_List.pdf}.
We acknowledge Pinar Demetci for her help in the discussion of genetic data pre-processing.

\section{Funding Statement}
This research received no specific grant from any funding agency in the public, commercial or not-for-profit sectors. 

\section{Competing Interests Statement}
The authors have no competing interests to declare.

\section{Contribution Statement}
All authors contributed to the design of the methodology and the experiments. M.G. implemented the data pre-processing, modeling, and data analysis. All authors discussed the results and contributed to the final manuscript.

\section{Data Availability}
The data underlying this article were provided by Alzheimer’s Disease Neuroimaging Initiative (ADNI) database (\url{https://adni.loni.usc.edu/}) by permission. Data will be shared on request to ADNI.

\printbibliography
%\bibliographystyle{vancouver}
%\bibliography{literature}

\clearpage
\newpage
\renewcommand{\thesection}{S\arabic{section}}
\renewcommand{\thefigure}{S\arabic{figure}}
\renewcommand{\thetable}{S\arabic{table}}
\setcounter{section}{0}
\setcounter{figure}{0}
\setcounter{table}{0}

\section*{MULTIMODAL ATTENTION-BASED DEEP LEARNING FOR ALZHEIMER'S DISEASE DIAGNOSIS}

\section*{Supplementary Material}
\label{section:supp-materials}

\section{Genetic Data Pre-processing}
\label{section:genetic}
The genetic data consists of the whole genome sequencing (WGS) data from 805 ADNI participants by Illumina’s non-Clinical Laboratory Improvement Amendments (non-CLIA) laboratory at roughly 30–40 × coverage in 2012 and 2013. The resulting variant call files (VCFs) have been generated by ADNI using Broad best practices (Burrows-Wheeler Aligner (BWA) and Genome Analysis Toolkit (GATK)-haplotype caller) in 2014. We first filtered the SNPs by the Hardy-Weinberg equilibrium (HWE) test for each site (p-values) by removing SNPs with HWE p $<$ 0.05. We then checked the genotype quality (GQ) and removed SNPs with GQ $<$ 20. Next, we filtered by minor allele frequency (MAF) and removed sites with MAF $<$ 0.01. Lastly, we performed genotype value filtering where we excluded sites based on the proportion of missing data and removed sites with a missing rate $>$ 0.05. After filtering with the above criteria, we utilized genes known to be related to Alzheimer’s Disease. In this step, we first downloaded a list of all AD-related genes from the AlzGene Database (\url{http://www.alzgene.org/}), which contains 680 genes in total. Then we searched these genes in the UCSC genome browser (\url{https://genome.ucsc.edu/}) and kept the 640 genes that matched NCBI RefSeq annotation. We extracted these gene regions from RefSeq Annotation (gff file) in Bed format and use them to filter the SNPs further. We only retain the genes that are located in these regions. After selecting the 680 genes of known association with AD, we had 547,863 SNPs left. As discussed in the {\hyperref[subsubsection:genetic_proc]{Genetic Data Pre-processing} Section}, we needed to find a way to reduce the number of features. We used a Random Forest Classifier to create a list of the most important features. Since this is a supervised method of creating features, this brought more promising results to the performance, in contrast to an approach such as principal component analysis (PCA) which is unsupervised. To find the best set of features, we tried using 50, 100, 150, and 200 forests as the parameter in the classifier. After creating the four sets of features, we used a validation set (10\% from the training) to do hyperparameter tuning (as described in \ref{section:hyper}), for each set of features, we found that the set from 100 forests performed the best on the validation set resulting in the accuracy described in {the \hyperref[subsection:uni_frame]{Performance of Unimodal Models} Section}. 

\section{Clinical Features}
\label{section:clinical}
\begin{figure}[H]
\begin{center}
    \caption{\textbf{Clinical feature importance.} The graph shows all the clinical features used in our model in order of most important to least important}
    \label{fig:clinical_imp}
  \includegraphics[scale=0.3]{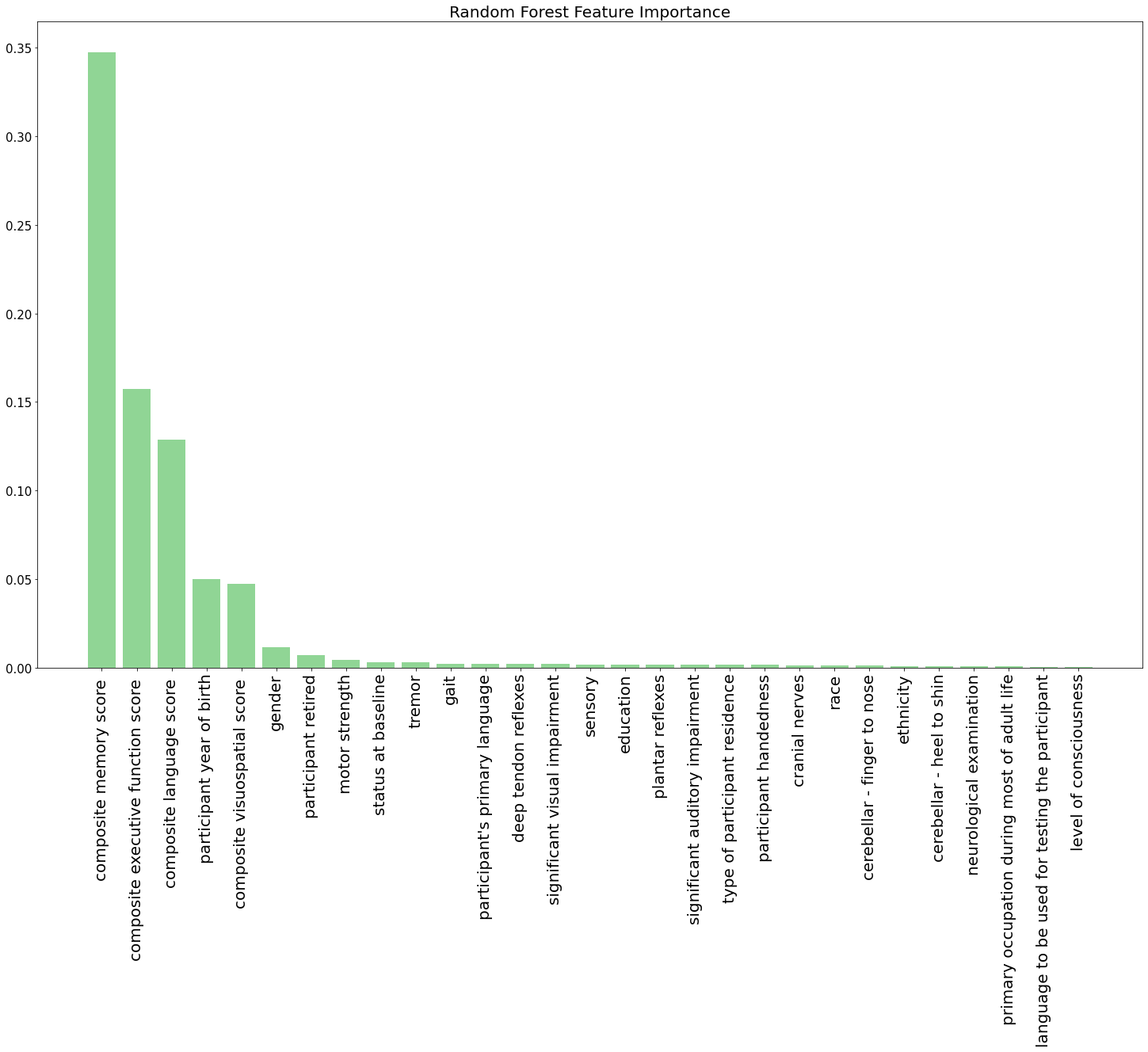}
  \end{center}
\end{figure}

In the {\hyperref[subsection:robust]{Model Robustness} Section} we discussed the value of clinical features to the model's performance. To ensure that there are no variables in the data that could potentially give an unfair advantage to the model (e.g.\ medication that a patient takes, only when they already have AD), we carefully examined all available variables. We fit a Random Forest classifier (from the scikit-learn package \cite{sklearn_api}) which outputs the features along with their importance score. The importance of a feature is computed as the (normalized) total reduction of the criterion brought by that feature \cite{sklearn_api}. Figure \ref{fig:clinical_imp} shows the full list of features and their importance. 

\newpage
\section{Imaging Data Pre-processing}
\label{section:img_preproc}
\begin{figure}[H]
\begin{center}
    \caption{\textbf{Validation F1-Score and Accuracy Trend as Number of Images Increases.} The graph shows that the unimodal imaging model does not significantly benefit from the addition of more images.}
    \label{fig:mri_exp}
  \includegraphics[scale=0.3]{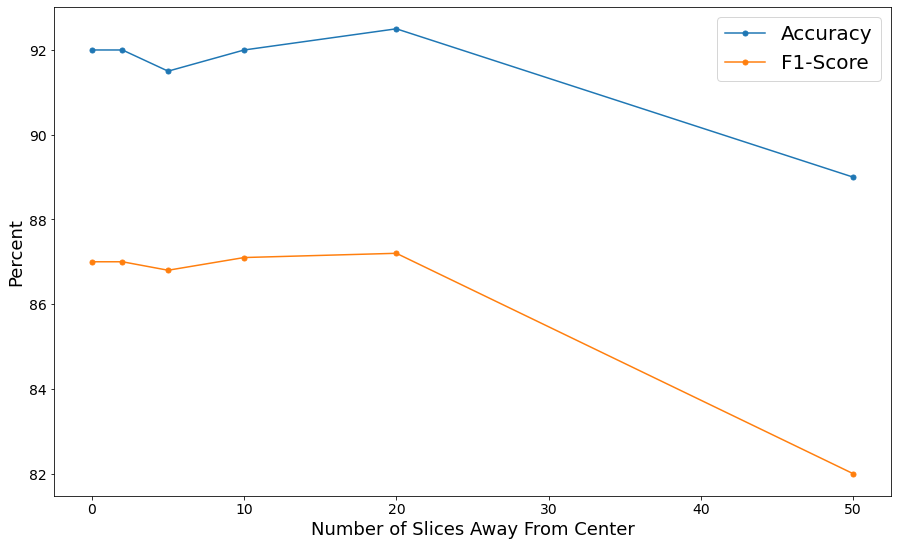}
  \end{center}
\end{figure}

The images used for our study are pre-processed by ADNI with specific image correction steps:
\begin{enumerate}
\item Gradwarp: gradwarp is a system-specific correction of image geometry distortion due to gradient non-linearity. The degree to which images are distorted due to gradient non-linearity varies with each specific gradient model. We anticipate that most users will prefer to use images which have been corrected for gradient non-linearity distortion in analyses.
\item B1 non-uniformity: this correction procedure employs the B1 calibration scans noted in the protocol above to correct the image intensity non-uniformity that results when RF transmission is performed with a more uniform body coil while reception is performed with a less uniform head coil.
\item N3: N3 is a histogram peak sharpening algorithm that is applied to all images. It is applied after grad warp and after B1 correction for systems on which these two correction steps are performed. N3 will reduce intensity non-uniformity due to the wave or the dielectric effect at 3T. 1.5T scans also undergo N3 processing to reduce residual intensity non-uniformity.
\end{enumerate}
We followed the same pre-processing steps as Bucholc et al. \cite{BUCHOLC2019157}, El-Sappagh et al. \cite{ELSAPPAGH2020197}, and Abuhmad et al. \cite{ABUHMED2021106688}, which relied on ADNI’s correction steps without further modification.
\newline 
To demonstrate that the performance of the unimodal imaging model is not significantly impacted by the addition of more brain slices, we report the optimized performance of the model with just the middle three slices, 2 more images per angle (6 more in total), 5 more images per angle, 10 more images per angle, 20 more images per angle, and 50 more images per angle. We report both F1-Scores and accuracy (averaged across 3-fold validation set), which follow a similar trend shown in \ref{fig:mri_exp}. The difference in performance between no additional slices (used in the paper) and 20 additional slices are all within 1 percent. When adding 50 slices to each angle, we observe a significant decline in performance. Thus, we proceeded with the original choice of just using the middle of the brain.  

\newpage
\noindent The decline in performance can be attributed to the fact that the slices further away from the center do not contain meaningful information and add noise to the model. The example below shows the middle three slices, followed by the outer 10 slices and outer most slices. 

\begin{figure}[H]
\begin{center}
    \caption{\textbf{Example of MRI slices as distance increases from the center.}}
    \label{fig:mri_slices}
  \includegraphics[scale=0.3]{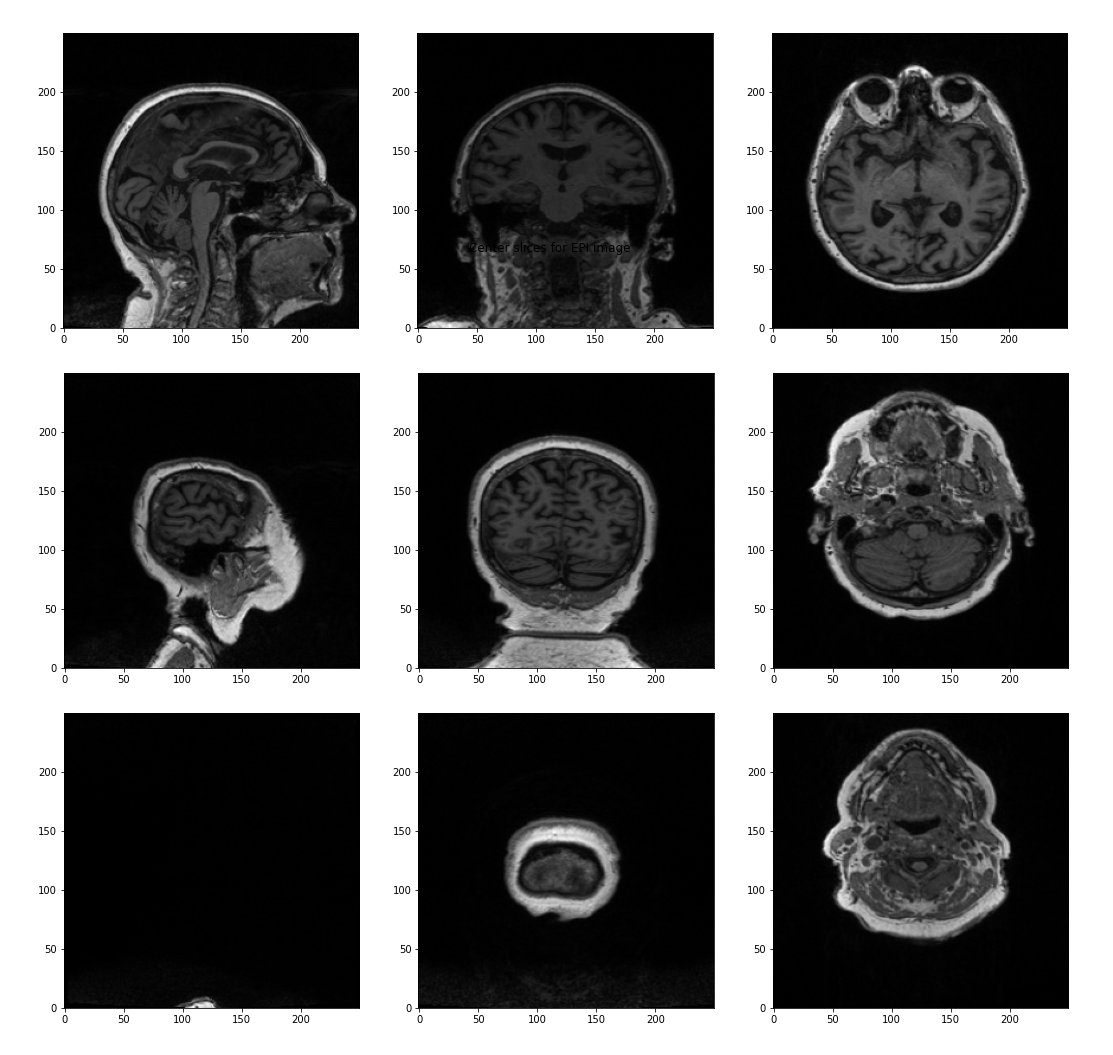}
  \end{center}
\end{figure}

\newpage
\section{Hyperparameter Tuning Methods}
\label{section:hyper}
To perform hyperparameter tuning for both unimodal and multimodal results, we randomly split the data into a training set (90\%) and a testing set (10\%). The testing set was set aside and withheld from tuning. We picked the best hyperparameters on the average validation accuracy of the 3-fold cross-validation scheme. For either a fully connected neural network or a convolutional neural network, the architecture, batch size, number of epochs, and learning rate were chosen via tuning. Table \ref{tab:hyper} describes all hyperparameters considered. 

\begin{table}[H]
\centering
\caption{\textbf{Hyperparameter Grid}}\label{tab:hyper}
\begin{tabular}{|l|l|}
\hline
\textbf{Hyperparameters} & \textbf{Values}                     \\ \hline
Learning Rate            & [0.00001, 0.1]                       \\ \hline
Dropout Values           & \{0.1, 0.2, 0.3, 0.4, 0.5\}      \\ \hline
Number of Layers         & [1, 6]                               \\ \hline
Batch Size               & \{16, 32, 64, 128\}               \\ \hline
Number of Epochs         & \{10, 20, 50, 80, 100, 150, 200\} \\ \hline
\end{tabular}
\end{table}

\noindent The hyperparameters that gave the highest accuracy for each type of model are shown in \ref{tab:best_hyper}. For the multimodal framework, the best unimodal neural network values were added into the architecture. 

\begin{table}[H]
\centering
\caption{\textbf{Best Hyperparameters}}\label{tab:best_hyper}
\begin{tabular}{|l|l|l|l|l|l|}
\hline
                  & \textbf{Learning Rate} & \textbf{Batch Size} & \textbf{Number of Layers} & \textbf{Dropout Value} & \textbf{Number of Epochs} \\ \hline
Unimodal Clinical & 0.0001                 & 32                  & 3                         & \{0.2, 0.3, 0.5\}      & 100                       \\ \hline
Unimodal Genetic  & 0.001                  & 32                  & 3                         & \{0.3,0.5\}            & 50                        \\ \hline
Unimodal Imaging  & 0.001                  & 32                  & 3                         & \{0.3,0.5\}            & 50                        \\ \hline
Multimodal        & 0.001                  & 32                  & \{3, 3, 3\}               & \{0.2, 0.3, 0.5\}      & 50                        \\ \hline

\end{tabular}
\end{table}

\section{Evaluation Metrics}
\label{eval_formula}
For our multi-class setting, we used the formulas below for each class. For example, for class 0 (control), we calculated the number of true positives, true negatives, false positives, and false negatives just for class 0. Then, we use “macro” averaged F1-score using the arithmetic mean of all the per-class F1-scores.

\begin{gather*}
Accuracy = \frac{TP+TN}{TP+TN+FP+FN}\\
Precision = \frac{TP}{TP+FP}\\
Recall = \frac{TP}{TP+FN}\\
F1 = \frac{2*Precision*Recall}{Precision+Recall} = \frac{2*TP}{2*TP+FP+FN}\\
\end{gather*}

\newpage
\section{Performance of Unimodal Models}
\label{eval_unimodal}
Table \ref{tab:uni} shows the numeric information presented in Figure \ref{fig:uni}. We report all four evaluation metrics for the best
neural network model for each modality - imaging, clinical, and genetic. The imaging model gives the best
performance overall, whereas the genetic modality gives the lowest performance. 

\begin{table}[H]
\caption{\textbf{Results of Unimodal Models.}}\label{tab:uni}
\centering
\begin{tabular}{|l|l|l|l|l|}
\hline
\textbf{} & \textbf{Accuracy (\%)} & \textbf{Precision (\%)} & \textbf{Recall (\%)} & \textbf{F1-Score (\%)} \\ \hline
Clinical  & 80.59             & 80.56              & 80.48           & 80.47             \\ \hline
Imaging   & 92.23             & 94.02              & 90.4            & 91.83             \\ \hline
Genetic   & 77.78             & 78.37              & 76.92           & 77.24             \\ \hline
\end{tabular}
\end{table}

\section{Model Robustness}
\label{tab_boxplot}
Table \ref{tab:attention} shows the numeric information presented in Figure \ref{fig:attention}.

\begin{table}[H]
\caption{\textbf{F1-Score Distribution for different attention-based and attention-free baselines.} The table shows the F1-score distribution across 100 random seeds to show the value of attention in a deep learning model. The table demonstrates that the combination of self-attention with cross-modal attention performs the best with the most narrow variation.} \label{tab:attention}
\scalebox{0.9}{%
\begin{tabular}{|l|l|l|l|l|l|}
\hline
& \textbf{Lower Whisker} & \textbf{Lower Quartile} & \textbf{Median} & \textbf{Upper Quartile} & \textbf{Upper Whisker} \\ \hline
Self-Att + Cross-Modal Att & 0.7767                 & 0.8268                  & 0.8799          & 0.9238                  & 1                      \\ \hline
Cross-Modal Att            & 0.4068                 & 0.6491                  & 0.7657          & 0.8268                  & 1                      \\ \hline
Self-Att                   & 0.7175                 & 0.8148                  & 0.8682          & 0.8799                  & 0.9238                 \\ \hline
No Attention               & 0.6396                 & 0.7714                  & 0.8148          & 0.8799                  & 0.9238                 \\ \hline
\end{tabular}}
\end{table}

\newpage
\section{Investigating Individual Class Performance}
\label{confusion}
We include confusion matrices for each of the 5 random initializations to supplement Table \ref{tab:classes} in the \hyperref[subsection:robust]{Model Robustness} Section. Each confusion matrix represents the results of our best multimodal model with respect to a random seed. 

\begin{figure}[H]

    \subfloat[Random Seed 1]{\includegraphics[width = 3in]{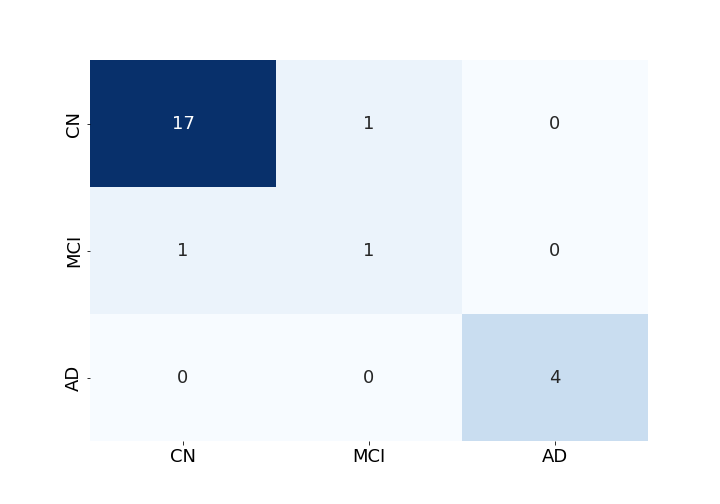}} 
    \subfloat[Random Seed 2]{\includegraphics[width = 3in]{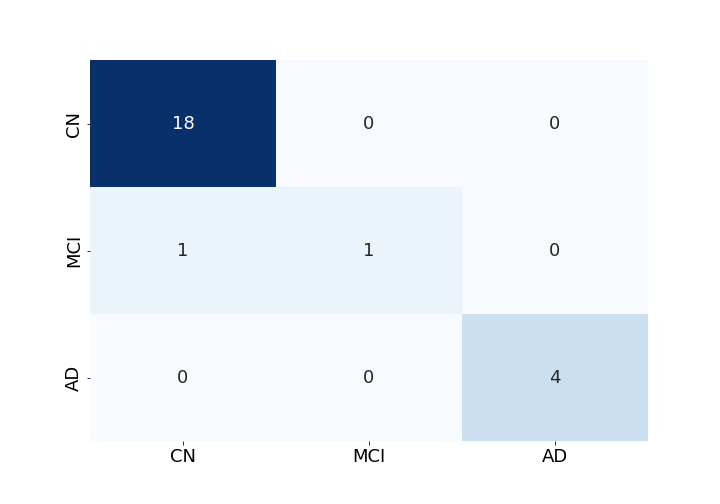}}\\
    \subfloat[Random Seed 3]{\includegraphics[width = 3in]{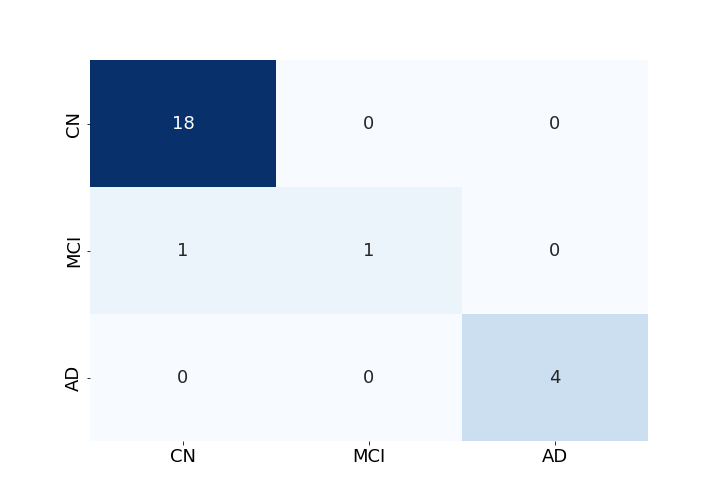}}
    \subfloat[Random Seed 4]{\includegraphics[width = 3in]{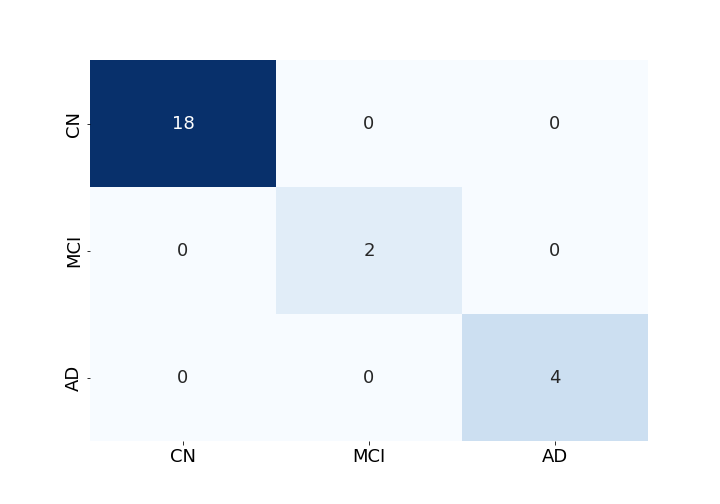}} \\
    \subfloat[Random Seed 5]{\includegraphics[width = 3in]{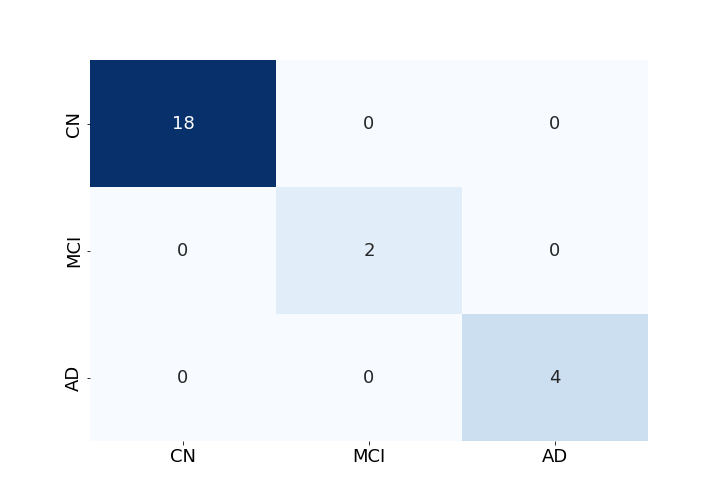}}
    \caption{\textbf{Confsion matricies for 5 random initializations of MADDi.}}
    \label{fig:confusion}

\end{figure}

\section{Sample Selection}
\label{section:samp_sel}
To demonstrate that our sample selection process was thorough, we show in Table \ref{tab:cross_val} the results of the models described in {\hyperref[subsection:robust]{Model Robustness} Section} on the 3-fold cross validation scheme. The metrics in the table are averaged across 5 random initializations. Since these results are similar to the ones reported on the test set in Table \ref{tab:comparison}, we consider the test set a fair sample of our data. 

\begin{table}[H]
\caption{\textbf{Cross-Validation Results}}\label{tab:cross_val}
\begin{tabular}{|l|l|l|l|}
\hline
                           & \textbf{F1-Score Val Set 1} & \textbf{F1-Score Val Set 2} & \textbf{F1-Score Val Set 3} \\ \hline
Cross-Modal Att + Self-Att & 89.74\%                     & 97.44\%                     & 92.30\%                     \\ \hline
Cross-Modal                & 87.18\%                     & 91.02\%                     & 88.46\%                     \\ \hline
Self-Att                   & 76.92\%                     & 85.89\%                     & 83.34\%                     \\ \hline
No Attention               & 79.48\%                     & 89.74\%                     & 87.17\%                     \\ \hline
\end{tabular}
\end{table}

\section{Evaluation of Modality Importance}
\label{section:modality_importance}
Table \ref{tab:modal_imp} shows the contribution and performance of each modality on the overlap patient set. The metrics were calculated as an average of five random initializations on a held-out test set. It captures the same information as Figure \ref{fig:eval_metric} in the {\hyperref[subsection:eval_imp]{Modality Importance} Section} but provides all the numeric results. 

\begin{table}[H]
\caption{\textbf{Evaluation of Modality Importance}}\label{tab:modal_imp}
\begin{tabular}{|l|l|l|l|l|}
\hline
                           & Accuracy (\%)     & Precision (\%)    & Recall (\%)       & F1-Score (\%)     \\ \hline
Clinical                   & 82.29\ { \scriptsize $\pm$ }  9.49\ & 78.92\ { \scriptsize $\pm$ } 3.68\  & 88.43\ { \scriptsize $\pm$ }  5.29\ & 78.30\ { \scriptsize $\pm$ }  6.70  \\ \hline
Genetic                    & 77.78\ { \scriptsize $\pm$ }  3.91\ & 78.37\ { \scriptsize $\pm$ }  4.64\ & 76.92\ { \scriptsize $\pm$ }  3.78\ & 77.24\ { \scriptsize $\pm$ }  4.03   \\ \hline
Imaging                    & 71.66\ { \scriptsize $\pm$ }  4.68   & 53.38\ { \scriptsize $\pm$ }  9.55\ & 62.03\ {\scriptsize $\pm$ }  9.77\ & 55.46\ { \scriptsize $\pm$ }  8.86\ \\ \hline
Clinical and Genetic       & 92.50\ { \scriptsize $\pm$ }  3.18\  & 87.05\ { \scriptsize $\pm$ }  9.36\ & 81.85\ { \scriptsize $\pm$ }  1.36\ & 83.21\ { \scriptsize $\pm$ }  4.21   \\ \hline
Genetic and Imaging        & 78.33\ { \scriptsize $\pm$ }  1.86\ & 50.88\ { \scriptsize $\pm$ }  8.19   & 52.59\ { \scriptsize $\pm$ }  6.88   & 50.07\ { \scriptsize $\pm$ }  4.28   \\ \hline
Imaging and Clinical       & 85.83\ { \scriptsize $\pm$ }  10.73  & 80.81\ { \scriptsize $\pm$ }  8.56   & 88.15\ { \scriptsize $\pm$ }  15.52  & 80.52\ { \scriptsize $\pm$ }  13.34  \\ \hline

\textbf{Clinical, Genetic, Imaging} & \textbf{96.88\ { \scriptsize $\pm$ }  3.33   }& \textbf{88.15\ { \scriptsize $\pm$ }  14.22}  & \textbf{91.23\ {\scriptsize $\pm$ }  13.37}  & \textbf{89.32\ { \scriptsize $\pm$ }  15.59 } \\ \hline
\end{tabular}
\end{table} 

\end{document}